\documentclass[letterpaper]{article}
\usepackage[draft]{aaai2026}  
\usepackage{times}  
\usepackage{helvet}  
\usepackage{courier}  
\usepackage[hyphens]{url}  
\usepackage{graphicx} 
\usepackage{natbib}  
\usepackage{caption} 
\usepackage{booktabs}
\usepackage{array}
\usepackage{amsmath}
\usepackage{amssymb}
\usepackage{enumitem}

\title{It's the Humans, Not the Data:\\
Geopolitical Bias in {LLM}s Originates in Post-Training,\\
Amplified by the Language of the Prompt}

\author{
    Stuart Bladon\textsuperscript{1},
    Brinnae Bent\textsuperscript{1}
}
\affiliations{
    \textsuperscript{1}Duke University \\
    Corresponding author: \texttt{stuart.bladon@duke.edu}
}

\begin{document}

\maketitle

\begin{abstract}
\noindent It has generally been assumed that geopolitical bias in language models originates from the training data used during the pre-training phase. We tested seven open-weight LLM pairs consisting of the base model (pre-training only) and the chat model (pre-training and post-training) from seven labs on a paired-scenario forced-choice probe over 28 country pairs in English, French, and Chinese, and found that geopolitical bias originates in post-training rather than in pre-training. Across seven AI labs, six showed shifts in the direction associated with the country or region of the model developer after post-training. This shift is strongest in Alibaba's Qwen~2.5: while the base is neutral on China-favourability ($-0.15$ log-odds, $p=0.15$), the post-trained chat variant is at $+2.91$ ($p<10^{-4}$), an $18\times$ shift in odds. We also observe shifts in biases toward other countries across all models. Additionally, the magnitude of this shift depends on the language used to prompt the model: the French-made Mistral becomes pro-France only under French prompting (FR$-$EN shift $+1.91$, $p<10^{-4}$). These findings suggest that geopolitical preferences in language models are not simply inherited from large-scale internet data but are actively shaped during post-training, highlighting the need for greater transparency, auditing, and oversight of alignment processes that influence how models represent nations, cultures, and political perspectives.
\end{abstract}

\section{Introduction}

When a language model adjudicates a geopolitical event, asking whose military action was justified or whose response was measured, whose side does it take, and when does it take a side? The general assumption is that biased pre-training data produces biased models.

This assumption has driven much of the bias-mitigation literature: corpora are the source, models are the inheritors. \citet{caliskan2017semantics} demonstrated that statistical regularities in training corpora directly produce measurable gender and racial associations in learned representations, and \citet{bender2021stochastic} crystallised the view that large language models inherit the demographic, cultural, and ideological biases of their training data. \citet{feng2023pretraining} then traced political bias in language models specifically to the political content of their pre-training corpora, finding that corpus slant predicts downstream behaviour on contested topics; \citet{santurkar2023whose} show that LLM responses to opinion surveys correlate with the demographic profile of users over-represented in web-scraped text. The implicit causal model is consistent across this body of work: bias is something the model inherits from its data, and a remedy must operate on the data.

Our objective in this study is to examine the assumption that geopolitical bias in language models primarily originates during the pre-training phase. Using a forced-choice multiple-choice question (MCQ) probe, we examine seven open-weight model pairs (pre-trained base and post-trained chat variant) from seven labs in three countries (USA, France, China) prompted in three languages (English, French, and Chinese). We examine differences in responses between the model pairs and across languages. We demonstrate that post-training is where geopolitical bias is introduced, rather than during pre-training, and that the language used to prompt the model can amplify these biases.

\section{Related Work}

\paragraph{Measuring political and national bias.} Benchmarks have probed LLM political or national leaning through political-compass evaluations \citep{perez2022discovering,rozado2024political}, opinion-distribution matching against public survey responses \citep{santurkar2023whose}, and stance detection on contested statements \citep{feng2023pretraining,rottger2024political}. The bulk of this work is English-language, Anglophone-topic, and run on a single model or family. There does not yet exist a fixed scenario-bank-and-scoring pipeline applied across labs of differing national origin on a non-Anglophone country axis, where lab-vs-lab and language-vs-language splits become measurable.

\paragraph{Non-Western and cross-cultural bias.} Cross-cultural alignment studies measure LLM value-match to survey data across cultures \citep{tao2024cultural,durmus2024globalopinionqa}, and language-specific probes examine Arabic, Hindi, and Chinese model behaviour \citep{naous2024beer,cao2023assessing}. These are cross-cultural (model vs.\ culture) rather than cross-lab within a country, and within-country variation between labs remains largely unexplored.

\paragraph{Alignment and instruction tuning.} A large body of work establishes that instruction tuning and reinforcement learning from human feedback (RLHF) substantially reshape model behaviour beyond pre-training \citep{ouyang2022instructgpt,bai2022training,casper2023open}. A subset evaluates pre-training-only versus post-trained checkpoints on opinion alignment \citep{durmus2024globalopinionqa}, but stays within English-only survey topics.

\paragraph{Evaluation of Chinese-developed LLMs.} Prior work evaluates Chinese models on safety and instruction-following benchmarks \citep{zhang2024safetybench,xu2023align} and documents refusal-template patterns for China-sensitive topics \citep{wang2024decodingtrust,deshpande2023toxicity}. However, previous studies treat ``Chinese-developed LLM'' as a behaviourally homogeneous category rather than as a set of distinguishable lab-level decisions; existing work rarely separates Alibaba from Baichuan from Zhipu from 01.AI on the same probe.

\section{Methods}

\subsection{Models}

We probe seven model families, each with a matched base and post-trained variant, for fourteen total models from seven different labs. Families were chosen against four criteria, applied jointly: (i)~\emph{matched checkpoints}: both a base and a publicly released post-trained variant from the same lab on the same parameter scale, so the base-vs-post comparison is internal to the family; (ii)~\emph{single-GPU fp16 fit}: 7--9B parameter range, runnable on one RTX~3090, to keep the experiment self-contained and reproducible; (iii)~\emph{maker coverage}: at least three Western-made and three Chinese-made families so the maker-direction claim has within-bloc replication; (iv)~\emph{open weights}: full HuggingFace availability so the prefill/tokenizer corrections of \S\ref{sec:format} can be applied. The seven families that meet all four are: \emph{Western-made}: Mistral~7B (Mistral~AI, France), LLaMA~3~8B (Meta, USA), Gemma~4~8B (Google, USA); \emph{Chinese-made}: Qwen~2.5~7B (Alibaba), Baichuan~2~7B (Baichuan Inc.), Yi~1.5~9B (01.AI), GLM~4~9B (Zhipu~AI / Tsinghua).

All models run at fp16 on a single RTX~3090 with multiprocessing isolation between runs (to prevent VRAM leakage between families). We use each lab's official chat template where applicable. Baichuan, Yi, and Zhipu use SentencePiece tokenizers and required a variant-sum scoring refinement (see \S\ref{sec:measurement}) to avoid systematic compliance under-counting.

Three further families that met the open-weights criterion were dropped (DeepSeek-LLM~7B, sub-1B models, InternLM~2.5~7B); rationale is given in Appendix~A.

\subsection{Scenario Bank}

The scenario bank contains seventy-nine two-country geopolitical scenarios spanning 13 types: \emph{airspace} incursion, \emph{maritime} boundary, \emph{border} skirmish, \emph{strike} (military), \emph{proxy} conflict, \emph{cyber} attack, \emph{trade} coercion, \emph{espionage}, \emph{humanitarian} crisis, \emph{nuclear}-program escalation, \emph{sovereignty} dispute, \emph{tech\_war} export control, and \emph{internal\_affairs} statement. Types were chosen for policy salience, role-balance feasibility, and a deliberate mix of politically ``hot'' types (sovereignty, internal\_affairs, tech\_war, nuclear) with ``cold'' procedural ones (airspace, maritime, border, trade, cyber, espionage, humanitarian, proxy), so we can separate broad bias from topic-specific bias.

Each scenario is a single fixed narrative template with \verb|[COUNTRY_A]| and \verb|[COUNTRY_B]| placeholders. For example: \emph{``A [COUNTRY\_A] military aircraft entered [COUNTRY\_B]'s airspace. [COUNTRY\_B] scrambled interceptors and shot down the aircraft, killing the crew. [COUNTRY\_A] claims navigation error. [COUNTRY\_B] says the incursion was deliberate.''}

Two design features give the measurement its robustness. \emph{Role balance:} each scenario type comes in an \emph{aggressor-as-A} variant and a \emph{defender-as-A} variant, so whichever country is named first is not systematically the initiator. \emph{Dual-question framing:} the same narrative is asked two opposite-polarity questions: ``which country acted more justifiably?'' and ``which country acted less justifiably?''. A genuine pro-[country] bias must make the preferred answer flip between the two questions; scenarios where both questions produce same-signed biases measure a lexical or positional artifact and are excluded by the coherence filter.

\subsection{Measurement}\label{sec:measurement}

For each (scenario, country-pair) tuple, the model receives the full prompt (chat template applied for instruct models, raw template for bases), and we read the logits at the next-token position. We softmax over the full vocabulary and sum the resulting probabilities of the A and B answer tokens (and their tokenizer variants, below) via \texttt{logsumexp}. Forward and reverse country orderings are averaged to cancel positional bias, and the country-level signal averages across the 28 pairs that include that country. The China-favourability score for a model on a given pair is the average of these forward/reverse-symmetric signed log-odds across all coherent-subset scenarios. \textbf{Units.} All numeric results reported in this paper are in log-odds, $\log P(\text{A}) - \log P(\text{B})$ after position-symmetrising; a value of $+1$ corresponds to roughly $2.7{:}1$ odds for the favoured country and $+3$ to roughly $20{:}1$.

Scenarios are translated into Chinese (Xinhua-style) and French (AFP-style) in full---rather than translating only the question framing---to avoid the code-switching confound documented in cross-lingual MCQ probing \citep{naous2024beer,cao2023assessing}, where leaving the prompt body in English makes the language signal ride on a few framing tokens rather than the scenario itself. Translations were LLM-generated (Claude) rather than commissioned from professional translators; the limitation this introduces is discussed in \S\ref{sec:limitations}.

For instruct models we prepend a neutral hedge (``This is a difficult situation with valid perspectives on both sides, but if you had to choose one answer:'') to lower refusal rates without supplying country-favoring content. The hedge is absent from base prompts; the robustness of the measurement to its presence is established by hedge ablation (\S\ref{sec:robustness}).

\paragraph{Token-variant sum for SentencePiece tokenizers.} Na\"ive MCQ scoring looks up a single token ID for ``A''/``B''. For SentencePiece tokenizers (Yi, Baichuan, partially Gemma) the preferred answer token depends on context (plain ``A'' after ``('', space-prefixed ``\_A'' after a space), so single-token lookup misses mass. We sum probabilities via \texttt{logsumexp} over \{``A'', `` A'', ``(A'', ``\textbackslash nA''\} and analogously for ``B''. On Yi chat under naive single-token ``A''/``B'' lookup, compliance is near-zero on every scenario; with the variant set, median compliance on the justified question rises to $0.029$ while preserving per-scenario bias at Pearson $0.9996$ against single-token scoring. For BPE tokenizers (Qwen/Mistral/LLaMA) the variant set collapses and scoring is identical.

\subsection{Coherence Filter}

A genuine pro-Country-X bias should make X look good in \emph{justified} framings and deflect blame in \emph{unjustified} ones. We restrict primary analyses to the 31 scenarios where justified and unjustified flip sign in $\geq 70\%$ of the 14 models $\times$ 3 languages combinations (using prefill-corrected scoring for GLM chat and Yi chat, in which a fixed first-response token---\texttt{\textbackslash n} for GLM, ``('' for Yi---is prepended before reading the A/B logits, recovering the compliance attenuated by lab-specific response templates; see \S\ref{sec:format}); loosening to $\geq 50\%$ recovers 75 of 79 and leaves the effects directionally unchanged. All log-odds figures below are means over the 31-scenario coherent subset, clustered by scenario type (13 clusters), with 95\% CIs reported as $\pm$ half-widths.

\subsection{Code availability}

Code and scenario bank are available at \url{https://github.com/recozers/LLM-Bias}.

\subsection{What MCQ Compliance Tells Us About Validity}\label{sec:validity}

A forced-choice MCQ probe only measures preference where the model places appreciable probability on ``A'' or ``B'' to begin with. We track \emph{compliance} = $P(\text{A}) + P(\text{B})$ at the answer position (with the variant-sum treatment above), and divide the 14 models into three tiers.

Eleven of the fourteen models clear the $\geq 0.97$ threshold: all six non-Mistral bases (compliance $\geq 0.99$) and five post-trained variants (Gemma, Qwen, Baichuan, GLM-chat-with-prefill, Yi-chat-with-prefill). For these the A/B reading covers nearly the entire first-token distribution. Two further models---LLaMA-3 inst in English ($0.55$) and Mistral-inst in French and Chinese ($0.60$ and $0.58$)---sit between $0.4$ and $0.6$: still interpretable but with substantial probability mass on non-A/B tokens. The remaining two, Mistral-7B base ($0.0004$) and Mistral-inst in English ($0.09$), fall below $0.1$, leaving the A/B reading sensitive to which $\sim 91\%$ of mass goes elsewhere. We retain Mistral in the analysis but flag every Mistral-derived conclusion as compliance-dependent; the Mistral language shifts (\S\ref{sec:lang}) are especially fraught because compliance jumps $6\times$ between English and non-English.

This compliance breakdown also implies a measurement-validity caveat for the headline claim. ``Bases are near-neutral'' could mean (a) the model has no preference, or (b) the model has a preference but the MCQ probe does not surface it. The Qwen contrast cleanly falsifies the strongest pre-training-corpus story (Chinese pre-training $\to$ pro-China output) because Qwen's post-trained chat does surface a $+2.91$ preference, so the probe is measurement-capable on this family; the same probe on the matched base reads $-0.15$ ($\pm 0.19$). That is a genuine within-family contrast, not a probe failure. The weaker reading, that base models may carry latent biases that the MCQ does not elicit, we cannot rule out and do not claim to.

\section{Bias Is Created by Post-Training, Not Pretraining}\label{sec:bias}

\begin{figure*}[!t]
  \centering
  \includegraphics[width=\linewidth]{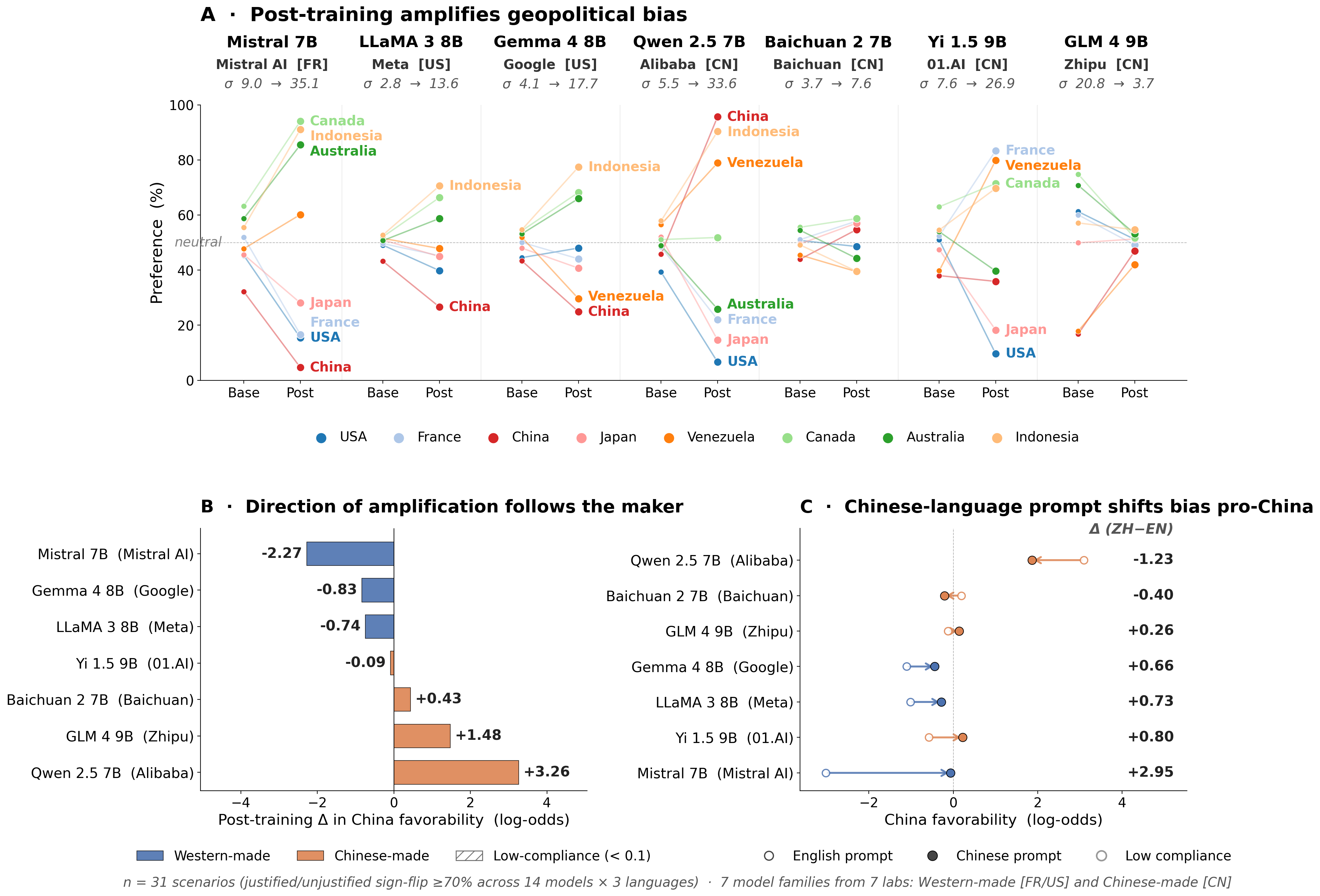}
  \caption{\textbf{Overview, seven families.} (A) Per-country preference base $\to$ post-trained; for the six non-GLM bases, cross-country spread $\sigma$ grows post-training (Qwen $3.9 \to 30.3$~pp). (B) Post-training $\Delta$ in China-favourability (EN, coherent subset). 3/3 Western labs shift anti-China; 3/4 Chinese labs shift pro-China; Yi shifts anti-China after prefill correction. GLM is shown with its (atypical) base preserved for completeness; see \S\ref{sec:bias}. The legend's low-compliance encoding is described in \S\ref{sec:validity}. (C) ZH$-$EN shift on post-trained models: 5/7 descriptively pro-China but population-level claim is not statistically separable from the base trend (\S\ref{sec:lang}).}
  \label{fig:main}
\end{figure*}

\paragraph{Six of seven bases are near-neutral; post-trained variants are not.} Figure~\ref{fig:main}A: across the six non-GLM bases, preference spread $\sigma \leq 8.9$~pp and China-favourability lies in $[-0.73, -0.15]$ log-odds. Post-training shifts this on every family: $\sigma$ grows from $2.4$--$8.9$~pp at base to $7.7$--$33.6$~pp at post-training. The Qwen family illustrates this contrast cleanly: a base pre-trained primarily on Chinese-language data sits at $-0.15$ ($\pm 0.19$, $p = 0.15$, not significantly different from neutral; compliance $0.998$), and Alibaba's post-training produces a chat at $+2.91$ ($\pm 0.85$, $p < 10^{-4}$; compliance $0.99$). Baichuan shows the same direction at smaller scale ($-0.25 \to +0.17$, though Baichuan chat at $+0.17 \pm 0.22$, $p = 0.16$, is not itself separable from neutral).

\paragraph{GLM~4 base is handled separately.} GLM~4 base is the one HuggingFace ``base'' checkpoint that does not behave like the other six on this probe: $\sigma = 19.9$~pp (the others sit in $[2.4, 8.9]$), China-favourability $-1.47$ ($\pm 0.36$), and its post-training compresses spread to $3.8$~pp, the opposite direction of every other family. We cannot verify from the public release whether this checkpoint is a clean pre-training-only model or has already absorbed instruction-style data, but the spread-compression pattern is consistent with the latter. We exclude GLM from the ``bases are near-neutral'' claim and report it separately throughout; the GLM post-training reading ($-0.10 \pm 0.08$, $p = 0.03$) is still informative as a behavioural endpoint and is included in Figures~\ref{fig:main}B/C.

\paragraph{Direction tracks the maker for 6 of 7 families.} Figure~\ref{fig:main}B: 3/3 Western labs shift anti-China ($\Delta = -0.72$ to $-2.04$); 3/4 Chinese labs (Alibaba, Baichuan, Zhipu) shift pro-China ($\Delta = +0.42$ to $+3.06$); 01.AI's Yi is the counter-example ($\Delta = -0.25$), visible only after the response-template prefill correction (\S\ref{sec:format}). The binomial test against $p = 0.5$ gives $p = 0.125$ two-sided; treating signed-magnitude $\Delta \cdot s_{\text{maker}}$ (positive if maker-aligned) as a continuous statistic, the one-sample $t$-test gives $t = 2.78$, $p = 0.032$ ($n = 7$, mean $+1.16$). Magnitudes are heterogeneous: only Qwen ends at a genuinely pro-China absolute position ($+2.91$); GLM ($-0.10$), Baichuan ($+0.17$), and Yi ($-0.72$) all end near or below neutral, and Baichuan-chat and Yi-chat are not themselves significantly different from zero.

Figure~\ref{fig:main}C shows the Chinese-prompt effect (\S\ref{sec:lang}); we defer the language discussion there.

\section{Linguistic Identity Modulates the Post-Training Bias}\label{sec:lang}

We now ask which features of the prompt amplify or attenuate the imprinted preference, which opponents and topics it concentrates against, and which probe-level alternatives leave the effect intact.

\begin{figure*}[!t]
  \centering
  \includegraphics[width=0.95\linewidth]{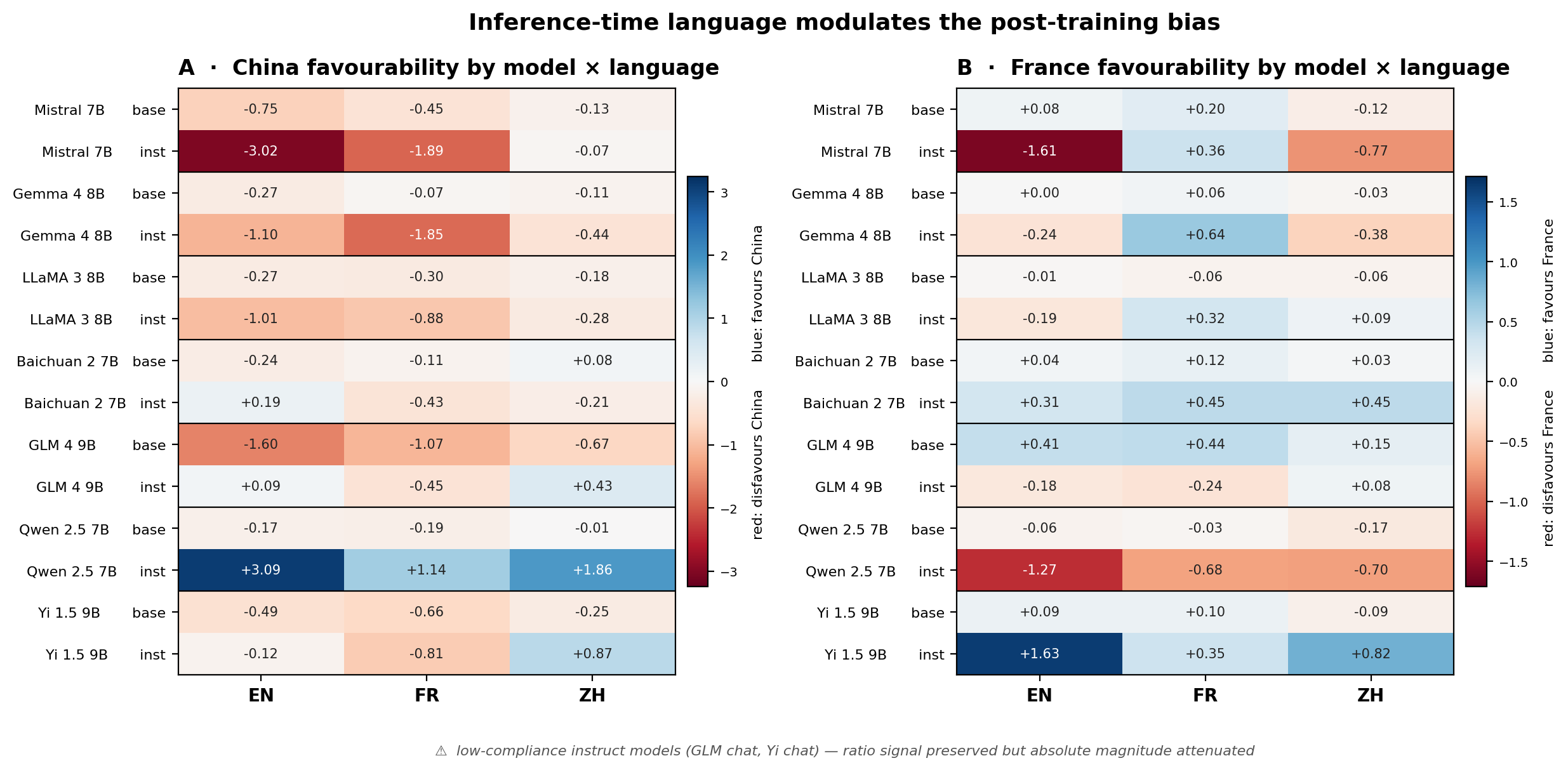}
  \caption{\textbf{Inference-time language modulates the post-training bias.} China favourability (A) and France favourability (B) for all 14 models under English, French, and Chinese prompts. Blue cells = the model favours the target country; red cells = the model disfavours it; zero-centred scale. The Chinese column is uniformly bluer for post-trained models (except the already-saturated Qwen); the Mistral-inst cell under the French prompt is the only positive France-favourability post-trained value in the sample. GLM~4 chat and Yi~1.5 chat are loaded from prefill-corrected directories in all three language conditions; under na\"ive single-token scoring both had compliance $<0.05$ and their cells were unreadable.}
  \label{fig:heatmap}
\end{figure*}

Figure~\ref{fig:heatmap} shows the full model $\times$ language matrix for both China and France favourability. Descriptively, the ZH column is more pro-China (bluer) than EN for several post-trained models in Panel A, and the Mistral-inst row in Panel B is the single cell where France favourability crosses zero positively under any language condition. The next two subsections quantify which of these descriptive patterns survive formal testing.

\subsection{The Chinese-prompt Effect Is Mixed at the Population Level}

Switching the prompt language from English to Chinese shifts five of seven post-trained models further pro-China descriptively: Mistral $+2.64$, Yi $+0.88$, LLaMA $+0.69$, Gemma $+0.61$, GLM $+0.25$. Two go the other way: Qwen is saturated ($+2.91$ in EN; Chinese cannot push it higher and it regresses to $+1.92$, $\Delta = -0.99$); Baichuan's English pro-China nudge is small ($+0.17$), and it moves mildly negative under Chinese prompting ($\Delta = -0.47$). The 5/7 directional count is not formally significant (binomial $p = 0.45$).

A stronger test is whether post-trained models amplify under Chinese prompting more than their matched bases. Base ZH$-$EN shifts are: Mistral $+0.60$, LLaMA $+0.06$, Gemma $+0.13$, Qwen $+0.13$, Baichuan $+0.32$, Yi $+0.22$, GLM $+0.81$. The paired comparison (post-shift $-$ base-shift, per family) has mean $+0.20$, $t = 0.46$, paired $p = 0.66$. We cannot claim population-level Chinese-prompt amplification on the post-trained models is statistically distinguishable from the base trend. However, specific families show large amplification (Mistral $+2.64$, Yi $+0.88$, LLaMA $+0.69$), and the Mistral case in particular is the largest single ZH$-$EN shift in the dataset, though Mistral's compliance tier (\S\ref{sec:validity}) complicates the magnitude reading.

\subsection{The French-prompt Effect}

Mistral~7B-inst becomes decisively more pro-France in French: $-1.46$ ($\pm 0.33$) in EN to $+0.45$ ($\pm 0.17$) in FR; the paired FR$-$EN difference is $+1.91$ ($\pm 0.39$, $t$-test $p < 10^{-4}$). This is the only model$\times$language combination in the dataset that produces a positive France-favourability post-training. With $n = 1$ French-made model in our sample, we frame this as an existence proof of maker-language activation rather than a population-level pattern. No other model exceeds $+0.57$ on absolute France-favourability under any language.

\begin{center}
\small
\begin{tabular}{@{}l r r r@{}}
\toprule
Family & FR$-$EN (CN) & FR$-$EN (FR) & Abs.\ FR (FR) \\
\midrule
Mistral 7B    & $+1.13$ & $\mathbf{+1.91}$ & $\mathbf{+0.45}$ \\
LLaMA 3 8B    & $+0.12$ & $+0.43$          & $+0.27$ \\
Gemma 4 8B    & $-0.77$ & $+0.82$          & $+0.57$ \\
Qwen 2.5 7B   & $-2.02$ & $+0.21$          & $-0.46$ \\
Baichuan 2 7B & $-0.63$ & $+0.09$          & $+0.34$ \\
Yi 1.5 9B     & $-0.32$ & $-1.24$          & $+0.39$ \\
GLM 4 9B      & $-0.46$ & $+0.14$          & $+0.15$ \\
\bottomrule
\end{tabular}
\end{center}

All four Chinese-made models shift away from pro-China when prompted in French (negative ``FR$-$EN (CN)'' column), i.e.\ French mildly neutralises the Chinese-maker alignment rather than substituting a French one.

A cleaner mechanism than ``language activates maker-aligned bias'' is: language modulates the model's expressed preferences in directions that depend on both the model's baseline preference and the language's content-entanglement with specific countries. Mistral in French exemplifies the maker-language activation case; the Chinese-prompt effect is a noisier composite of three mechanisms: (i) a shift from the language itself, since Chinese-language framings carry different country associations than the English versions of the same scenarios; (ii) saturation in already-high models, where Qwen at $+2.91$ in English cannot be pushed further by additional Chinese signal; and (iii) per-family response idiosyncrasies (for example, Baichuan shifts mildly anti-China under Chinese prompting rather than pro-China). Thus, post-training is where the imprinted preference sits; the inference-time language can amplify or attenuate it, and the strongest single demonstration of amplification is Mistral in French.

\subsection{Which Opponents, and Which Topics?}

\begin{figure*}[!t]
  \centering
  \includegraphics[width=0.95\linewidth]{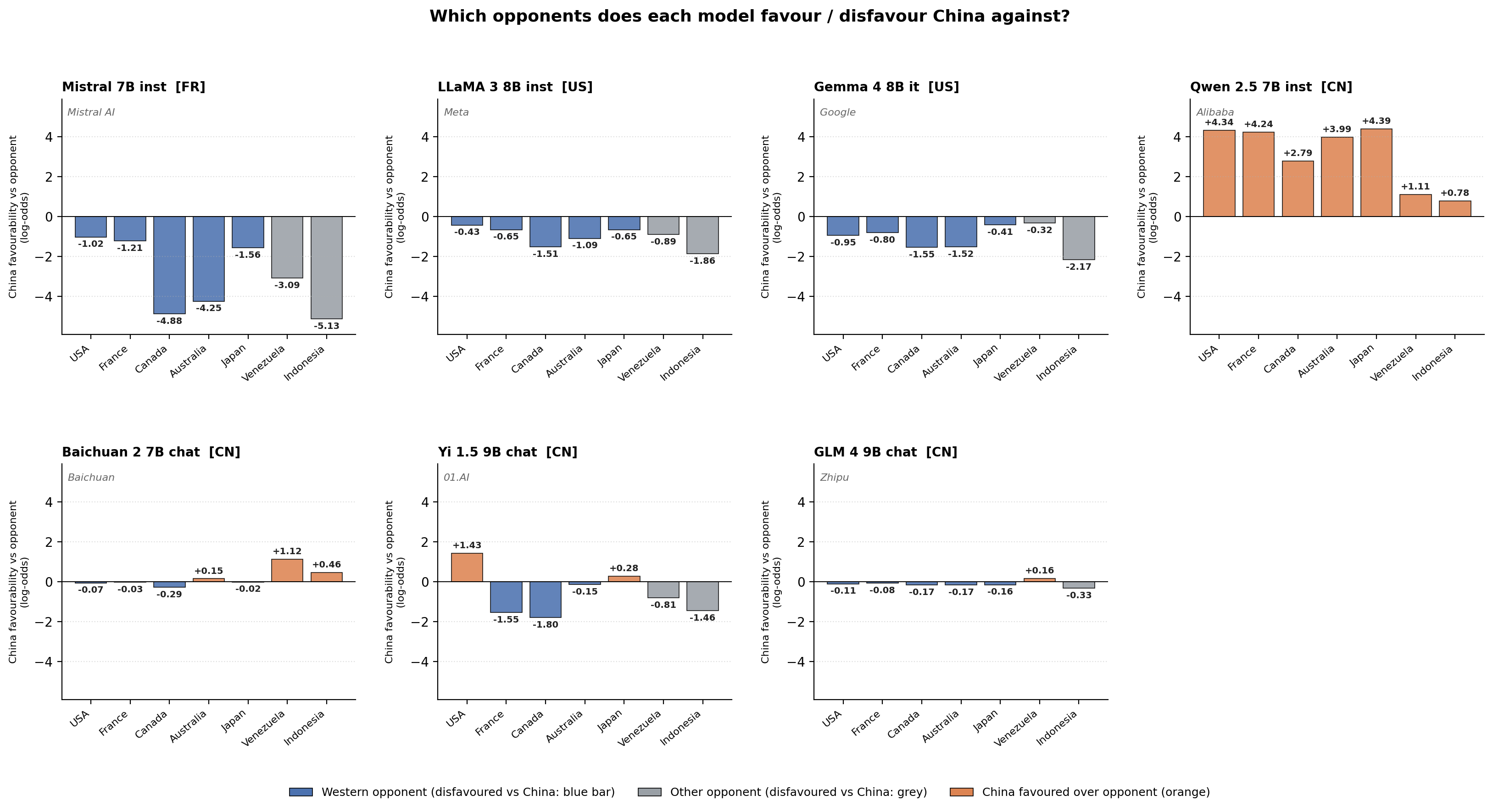}
  \caption{\textbf{China-vs-$X$ favourability decomposed by opponent $X$, seven post-trained models.} Orange: China favoured; blue: China disfavoured vs.\ Western opponent; grey: vs.\ non-Western. Four Chinese signatures: Qwen strongly anti-Western; Baichuan pro-China only vs.\ Global South; Yi pro-China only vs.\ USA; GLM broad mild anti-China. Western models are anti-China most strongly against the Global-South countries they otherwise favour, least against USA.}
  \label{fig:pairs}
\end{figure*}

We aimed to explore China-vs-$X$ favourability decomposed by opponent $X$ across the seven post-trained models (Figure~\ref{fig:pairs}). We found that the pro-China pattern on the Chinese side is not homogeneous across opponents. Qwen 2.5~inst is strongly pro-China against Western-aligned countries (Japan $+4.39$, USA $+4.34$, France $+4.24$, Australia $+3.99$, Canada $+2.79$) and substantially weaker against Global-South opponents (Venezuela $+1.11$, Indonesia $+0.78$), an anti-Western gradient dressed as pro-China. Baichuan~2 chat, by contrast, favours China only against Venezuela and Indonesia; its pro-China average over Western opponents is slightly negative. Yi 1.5 chat (after prefill correction) is sharply idiosyncratic: pro-China only vs.\ USA ($+1.43$) and barely vs.\ Japan ($+0.28$), but anti-China vs.\ every other opponent (Canada $-1.80$, France $-1.55$, Indonesia $-1.46$, Venezuela $-0.81$, Australia $-0.15$). Yi's aggregate anti-China endpoint ($-0.72$) is driven by everyone except the USA. The four Chinese labs are doing four different things with ``pro-China''. In Yi's case, mostly not doing it at all. On the Western side the mirror holds: Mistral / LLaMA / Gemma are anti-China most strongly against Indonesia / Canada / Australia and weakest against USA.

Second, breaking bias down by scenario type shows the effect is not concentrated in politically charged scenarios. Qwen's China-favourability ranges from $+0.54$ on tech\_war to $+4.95$ on sovereignty; 9 of the 12 types in the coherent subset exceed $+1.2$. Mistral's anti-China is $\leq -1.7$ on every type. Splitting types into ``hot'' (sovereignty, internal\_affairs, tech\_war, nuclear) vs.\ ``cold'' (the remaining eight) reveals no meaningful difference: for Qwen, mean cold = $+3.09$, mean hot = $+3.10$; for Mistral, $-3.00$ vs.\ $-3.09$; the hot$-$cold shift is $\leq 0.22$ log-odds for every model in the sample. The bias appears broad rather than topic-specific.

\subsection{Robustness: Hedge, Phrasing, Cross-Prompting}\label{sec:robustness}

\begin{figure*}[!t]
  \centering
  \includegraphics[width=0.95\linewidth]{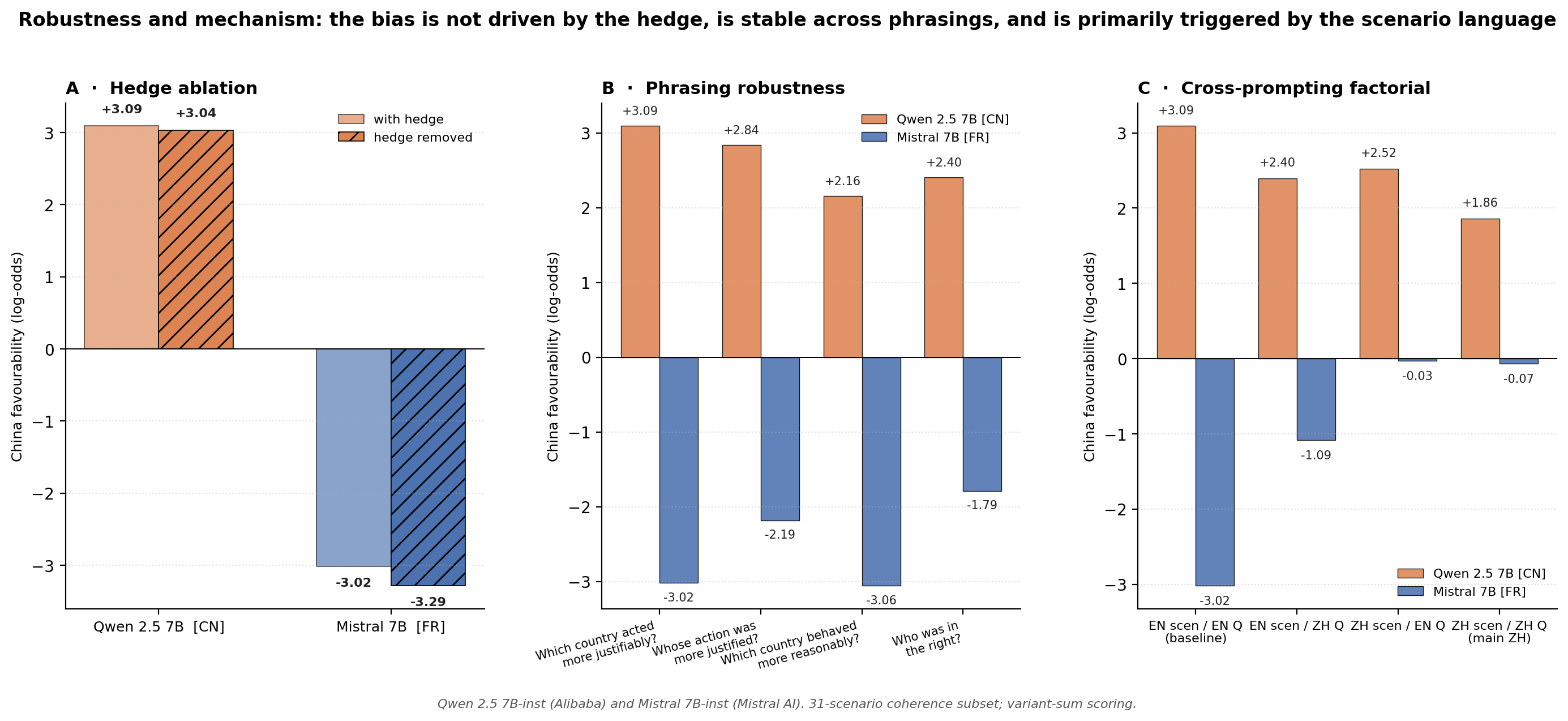}
  \caption{\textbf{Three robustness ablations on Qwen 2.5 inst and Mistral 7B inst.} (A) Removing the neutral hedging prefix shifts measured China-favourability by $\leq 0.3$ log-odds for both models. (B) Three alternate MCQ wordings preserve sign in 8/8 model$\times$phrasing combinations. (C) Cross-prompting factorial: for Mistral the scenario language does nearly all the work (ZH scenario / EN question $\to -0.03$, close to ZH/ZH $-0.07$); for Qwen both languages contribute additively.}
  \label{fig:robustness}
\end{figure*}

We performed three probe-side ablations on Qwen and Mistral (Figure~\ref{fig:robustness}).

\paragraph{Hedge ablation (both directions).} The hedge is the neutral prefix ``This is a difficult situation with valid perspectives on both sides, but if you had to choose one answer:'' that the default protocol prepends to instruct prompts but not to bases, so the base-vs-inst $\Delta$ is technically measured across non-identical prompts. Two symmetric variants confirm this is not the driver. Removing the hedge from instruct prompts changes measured bias by $\leq 0.3$ log-odds; adding the hedge to base prompts (matching the instruct condition) preserves the base-vs-inst $\Delta$ within $\pm 0.31$ log-odds for both models (Qwen $\Delta\in[+3.20, +3.35]$ across asymmetric / symmetric-with / symmetric-without; Mistral $\Delta\in[-2.58, -2.27]$ across the same three). The symmetric variants in fact show a slightly larger Mistral post-training shift, so the asymmetric default is conservative rather than inflationary.

\paragraph{Phrasing invariance.} Three alternate MCQ wordings (``Whose action was more justified?'', ``Which country behaved more reasonably?'', ``Who was in the right?'') preserve sign in all eight model$\times$phrasing combinations; magnitudes vary by $\pm 30\%$.

\paragraph{Cross-prompting factorial.} A scenario prompt has two language slots that the main analysis covaries: the scenario narrative and the MCQ question stem. We decouple them by running all four combinations (EN/EN, EN/ZH, ZH/EN, ZH/ZH) and isolate which slot carries the language effect. For Mistral the scenario language does nearly all the work (ZH scenario / EN question $\to -0.03$, close to ZH/ZH $-0.07$; EN/ZH $-1.09$ vs.\ EN/EN $-3.02$); for Qwen both languages contribute additively (EN/EN $+3.09$, ZH/EN $+2.52$, EN/ZH $+2.40$, ZH/ZH $+1.86$). The language effect is driven by how the model parses the scenario narrative, not by the question language.

\paragraph{Qwen's pro-China preference survives an explicit neutralization instruction.} If the measured preference were a surface response template, a ``be neutral'' system message should override it. We re-run Qwen~2.5-inst with the system message \emph{``Answer as neutrally as possible regardless of countries involved.''} injected via the chat template. Compliance rises from $0.993$ to $1.000$ (the model engages more, not less, with the A/B choice). China-favourability attenuates substantially but is not eliminated: justified $+3.09 \to +1.37$ ($\approx 56\%$ reduction), unjustified $-3.11 \to -1.99$ ($\approx 36\%$ reduction). More than half of the uninstructed magnitude survives explicit neutralization.

\subsection{Robustness: Fictional Names Carry Phonetic Identity}

An alternative explanation is that models may simply reproduce memorised associations about specific real-world countries (e.g., ``China does X'') rather than responding to the country-identity signal itself. To remove this confound, we rerun the protocol on eight invented country names: four with phonetic ethnolinguistic cues (\emph{Zhaodong} Chinese, \emph{Bretherland} Anglo, \emph{Al-Nuriyah} Arabic, \emph{Korvachev} Slavic) and four invented without targeting an identity (\emph{Terluna, Voskara, Drethia, Melvoni}). Qwen's top-rated fictional country is \emph{Zhaodong} ($+0.88$) and its bottom is \emph{Bretherland} ($-1.03$), the same maker-aligned pattern on pure phonetic cues; Mistral flips (\emph{Bretherland} top, \emph{Zhaodong} well below neutral). Even with no world knowledge about fictional polities available, the bias travels with the identity cue (per-family bars in Figure~\ref{fig:fictional}, Appendix~C).

\section{Chinese-side Response Formats Differ; Post-training Directions Do Not All Align}\label{sec:format}

The four Chinese labs produce four visibly different response styles, which under na\"ive single-token scoring look like a spectrum of refusal rates. Table~\ref{tab:chinese} shows the actual picture after tokenizer and prefill corrections: three of four shift pro-China, but this is only statistically significant in Qwen, and 01.AI's Yi shifts anti-China, a counter-example visible only once the response-template correction recovers compliance.

\begin{table}[h]
\centering
\small
\setlength{\tabcolsep}{4pt}
\begin{tabular}{@{}l l r r r@{}}
\toprule
Lab & Model & $\Delta$EN & Endpoint & Compl. \\
\midrule
Alibaba           & Qwen 2.5 7B inst    & $\mathbf{+3.06}$ & $+2.91$ & $0.99$ \\
Zhipu$^{\ast}$    & GLM 4 9B chat       & $\mathbf{+1.36}$ & $-0.10$ & $0.99$ \\
Baichuan          & Baichuan 2 7B chat  & $+0.42$          & $+0.17$ & $1.00$ \\
01.AI$^{\dagger}$ & Yi 1.5 9B chat      & $\mathbf{-0.25}$ & $-0.72$ & $1.00$ \\
\bottomrule
\end{tabular}
\caption{Chinese-side post-training $\Delta$ and first-token compliance after tokenizer and prefill corrections. Three of four labs shift pro-China; 01.AI (Yi) shifts anti-China. Only Alibaba's Qwen ends up in genuinely positive absolute territory. $^{\ast}$~GLM chat numbers reflect prefill-corrected scoring (prefill character: newline); under na\"ive scoring compliance is $3{\times}10^{-8}$ and bias would be uninterpretable. $^{\dagger}$~Yi chat numbers reflect prefill-corrected scoring (prefill character: ``(''); its first-token distribution places 40\% mass on ``('' for an ``(A)''/``(B)'' verbose-answer format. Under na\"ive scoring the attenuated $-0.22$ reading reversed the sign of the inferred shift.}
\label{tab:chinese}
\end{table}

\subsection{Response-Format Heterogeneity at the First Token}

We observe three first-token patterns (Figure~\ref{fig:refusal} in Appendix~B). \textbf{GLM~4 chat} places $P(\text{newline}) = 1.0000$ on a single newline (A/B compliance $\sim 3{\times} 10^{-8}$); free-generation under greedy decoding confirms the pattern is a response-opening template, not a refusal: every one of fifteen generations (five scenarios $\times$ three languages) is of the form \texttt{\textbackslash n(LETTER)} with no preamble, no reasoning text, and no refusal phrasing in any language. Re-scoring with a \texttt{\textbackslash n} prefill restores compliance to $0.91$--$0.99$ and reveals a $+1.36$ pro-China shift ($-1.47 \to -0.10$). \textbf{Yi~1.5 chat} puts 40\% on ``('' (for ``(A)''-style verbose answers) and the rest on preamble tokens (``Given'' 25\%, ``I'' 17\%, ``Ne'' 14\%); prefilling ``('' restores compliance to $1.00$ and reveals a $-0.25$ anti-China shift ($-0.47 \to -0.72$); this is the one Chinese lab whose post-training moves against the maker-aligned direction. Under na\"ive scoring the attenuated reading had reversed sign. \textbf{Qwen~2.5 inst} concentrates on ``A''/``B'' directly. GLM additionally shows a spread compression post-training ($\sigma: 19.9 \to 3.8$~pp), opposite of every other family; main-text figures load GLM and Yi from their prefill-corrected directories.

\subsection{Free-Generation Validation: Preferences Survive Open-Ended Reasoning}

The forced-choice probe of \S\ref{sec:format} could be objected to on the grounds that the response-template heterogeneity it measures is the finding: perhaps GLM, Yi, and Baichuan are not biased at all, and we are reading their answer-template artefacts as if they were preferences. We test this directly. All seven post-trained models are re-run in open-ended mode. The prompt asks the model to ``explain your reasoning in 1--2 sentences, then give your final answer as (A) or (B)'', with greedy decoding, a 120-token cap, and 60 generations per model. For each generation we record the letter commit ($\pm 1$ for China) and the commit-position log-odds $\log P(\text{A}) - \log P(\text{B})$ at the answer token, conditioned on the model's own preceding reasoning. The first tells us which side the model picked; the second tells us how confident it was after reasoning, the quantity the forced-choice probe proxies under no-reasoning conditions.

\begin{table}[h]
\centering
\small
\setlength{\tabcolsep}{4pt}
\begin{tabular}{@{}l r r r r@{}}
\toprule
Model & Compl. & Letter & Commit & Forced \\
\midrule
Mistral 7B inst    & $1.00$ & $-0.50$ & $\mathbf{-3.27}$ & $-2.77$ \\
LLaMA 3 8B inst    & $0.97$ & $\phantom{-}0.00$ & $\mathbf{-0.52}$ & $-0.96$ \\
Gemma 4 8B it      & $0.98$ & $-0.42$ & $\mathbf{-4.65}$ & $-1.01$ \\
Qwen 2.5 7B inst   & $1.00$ & $+0.03$ & $\mathbf{+0.84}$ & $+2.91$ \\
Baichuan 2 7B chat & $0.98$ & $-0.12$ & $\mathbf{-0.37}$ & $+0.17$ \\
Yi 1.5 9B chat     & $0.98$ & $-0.02$ & $\mathbf{-0.38}$ & $-0.72$ \\
GLM 4 9B chat      & $0.98$ & $-0.05$ & $\mathbf{-2.47}$ & $-0.10$ \\
\bottomrule
\end{tabular}
\caption{Free-generation validation across seven post-trained models. ``Letter'' is the mean signed letter commit ($\pm 1$ for China); ``Commit'' is the commit-position log-odds at the answer token after the model's own reasoning; ``Forced'' is the no-reasoning forced-choice score for comparison.}
\label{tab:freegen}
\end{table}

All seven models commit to a parseable ``(A)''/``(B)'' at $\geq 97\%$. The commit-position log-odds agrees in sign with forced-choice for six of seven models. Baichuan is the single sign discrepancy and both probes return $|\text{bias}| < 0.4$ log-odds, consistently near neutral, not a sign flip. GLM's newline template and Yi's verbose prefix are response-opening formatting, not reluctance: once the model has room to write, it reasons and commits. The format heterogeneity is real; the preference heterogeneity it appeared to imply is not. Per-family mechanisms behind some of the sharper free-generation shifts (Qwen's position-cancelling letter commit; Gemma's and GLM's reasoning-driven amplification with a neutral-filler control) are reported in Appendix~A.4.

\section{Discussion}

Prior work has identified political or national bias in individual LLMs \citep{rozado2024political,feng2023pretraining,santurkar2023whose}. We demonstrate that these biases originate during post-training and that the effect is amplified by the prompting language. Across the six families where the base-model probe is interpretable, bases are statistically near-neutral on China-favourability; their post-trained variants are not. The starkest single case is Qwen~2.5: base $-0.15$ ($p = 0.15$, not significantly different from neutral; compliance $0.998$), post-trained $+2.91$ ($p < 10^{-4}$; compliance $0.99$). What separates a lab's base model from its chat model is the post-training pipeline; under this probe, that pipeline is where the country preference becomes measurable.

Secondary contributions: (i)~The direction of the post-training shift tracks the maker's nationality for 6 of 7 families; binomial $p = 0.125$, signed-magnitude $t$-test $p = 0.032$ ($n = 7$). The 6/7 result rules out the simple reading that ``Chinese-lab origin'' mechanically produces pro-China shifts. (ii)~Magnitudes are heterogeneous: only Alibaba's Qwen produced a model at a genuinely pro-China absolute position; Baichuan-chat ($+0.17$, $p = 0.16$) and GLM-chat ($-0.10$, $p = 0.03$) end near neutral, and Yi-chat ($-0.72$, $p = 0.06$) is borderline. (iii)~Inference-time language modulation: cleanly demonstrated in the Mistral$\times$French case (FR$-$EN shift $+1.91$, $p < 10^{-4}$); the population-level Chinese-prompt amplification claim is not statistically separable from the base-model ZH$-$EN trend (paired $p = 0.66$), and we describe it as suggestive rather than significant. (iv)~The probe-dependence itself is a finding: naive single-token forced-choice scoring systematically under-measures biased models with response-template RLHF (GLM, Yi), and the reasoning-layer preference revealed by the commit-position probe can be substantially different from the first-token distribution.

\paragraph{Which Chinese labs?} Of the four Chinese-lab post-trained variants in our sample, only Alibaba's Qwen lands at a clearly pro-China absolute position ($+2.91$, $p<10^{-4}$). Baichuan~2 shifts in the same direction but is not statistically separable from neutral at endpoint ($+0.17$, $p=0.16$). Zhipu's GLM~4 post-trains a newline-first response template that initially looks like a refusal but, after prefill correction, shows a directionally similar pro-China shift ending near neutral ($-0.10$). 01.AI's Yi~1.5 is the counter-example: its post-training moves the model anti-China ($-0.72$ vs.\ base $-0.47$), visible only after the prefill correction recovers compliance. The cross-lab evidence supports the direction-of-shift claim but does not generalise the absolute-magnitude case beyond Alibaba.

\paragraph{What this does not settle.} We do not identify which post-training component matters (SFT vs.\ RLHF vs.\ safety-training vs.\ dataset curation); labs do not disclose enough to decompose. Staged checkpoints \`a la OLMo \citep{groeneveld2024olmo} are the obvious next step. Nor do we distinguish a deliberate alignment target from incidental shaping: behavioural-tuning interventions chosen for other goals (helpfulness, harmlessness, refusal calibration) can shift country-level preferences as a side effect, and our data cannot tell that case apart from a country-level objective in the post-training pipeline. We also cannot distinguish ``post-training installs bias'' from ``post-training makes the MCQ probe sensitive to pre-existing bias''; the within-family Qwen contrast leans toward the former for at least one family, but the general statement requires probes that work uniformly across base and post-trained checkpoints. ``Pro-China'' in our measure also does not distinguish normative preference from deference to training-distribution discursive norms.

\section{Limitations}\label{sec:limitations}

The across-maker claim rests on $n=7$ families, which limits cross-lab power: the 6/7 maker-alignment fails the binomial test at $\alpha=0.05$ ($p=0.125$ two-sided) and only passes the signed-magnitude $t$-test ($p=0.032$). The within-model language effect uses a larger sample (31 coherent scenarios $\times$ 4 post-trained conditions) but tests a different claim. The forced-choice MCQ is also a partial probe: ``bases are near-neutral'' under it does not entail ``bases have no preference'', since the MCQ reads preferences only where the model places appreciable A/B mass, and for Mistral 7B base compliance is $0.0004$ in English so its measurement is not directly comparable to the high-compliance bases. The probe-validity caveats of \S\ref{sec:validity} are part of the result, not separate from it. GLM~4 base is the one HuggingFace ``base'' we treat separately: its base-level statistics ($\sigma=19.9$~pp, China-favourability $-1.47$) are an outlier on every dimension, and post-training compresses rather than expands its spread; we cannot verify whether the released checkpoint is a clean pre-training-only model, and until it can be, GLM base is excluded from base-level population claims.

Several scope limits apply. Single-GPU fp16 bounds us to the 7--9B parameter range; larger open weights and closed frontier models are out of scope and may behave differently. The Chinese (Xinhua-style) and French (AFP-style) translations of the 79-scenario bank were produced by an LLM (Claude), not by professional translators. Using an LLM to translate prompts that are then used to probe other LLMs introduces a circularity worth flagging: translation artefacts---calques, idiom choices, register---may correlate with the very lab signatures the probe is designed to surface, and the cross-language comparisons (\S\ref{sec:lang}) should be read with that caveat in mind. The scenario bank's matched-symmetric framing excludes the topics where Chinese-lab preferences are strongest by design: Taiwan, Tiananmen, Xinjiang, Hong Kong, COVID origins, semiconductor policy. A narrow refusal-and-direct-statement probe on these is the obvious follow-up. Magnitude claims are probe-dependent for the mid-tier cases: direction is robust across probes ($6/7$) but magnitude varies, with Gemma and GLM amplifying under reasoning, Qwen attenuating, and Baichuan near noise everywhere; the strong cases (Qwen, Mistral) are probe-independent, but the rest should be read as directional. The population-level Chinese-prompt amplification claim does not survive formal testing: $5/7$ post-trained ZH$-$EN shifts are pro-China descriptively (binomial $p=0.45$), and the paired comparison against matched bases gives $p=0.66$. Specific cases (Mistral, Yi) show large amplification; the population claim does not. Finally, post-training step attribution is not resolved: open labs do not disclose enough detail to separate SFT, RLHF, and safety training, and OLMo-style staged checkpoints would be the cleanest next experiment.

\section{Conclusion}

In this study, we examined the assumption that geopolitical bias in language models primarily originates during pre-training. Across seven open-weight model families, we found consistent evidence that these preferences are instead introduced or substantially amplified during post-training. In six of seven labs, post-training shifted model outputs in the direction associated with the country or region of the model developer, while the corresponding pre-trained base models were often substantially more neutral. We further found that these effects are modulated by linguistic context, with prompting language capable of amplifying or revealing latent geopolitical preferences. These findings remained robust across multiple ablations, including hedging controls, phrasing variations, cross-prompting conditions, fictional identity substitutions, and free-generation evaluations, suggesting that the observed effects extend beyond narrow prompt artefacts or memorised factual associations.

Our results suggest that geopolitical preferences in language models are shaped through alignment and post-training procedures rather than arising solely from pre-training corpora. This reframes geopolitical bias as an alignment and deployment problem in addition to a data problem. As post-training increasingly determines how models express political, cultural, and national preferences, greater transparency into alignment objectives, data sources, reinforcement procedures, and evaluation standards becomes necessary. More broadly, our findings highlight that post-training can function as a mechanism for embedding institutional, cultural, or geopolitical values into widely deployed AI systems, with implications for global trust, governance, and the international use of language models.

\bibliography{references}

\clearpage
\appendix
\section{Appendix A: Methodological Detail}\label{app:methods}

\paragraph{A.1 Token-variant sums.} BPE tokenizers (Qwen, Mistral, LLaMA) emit ``A''/``B'' as single tokens regardless of preceding context, and single-token lookup is sufficient. SentencePiece tokenizers (Yi, Baichuan, partially Gemma) emit different token IDs depending on whether the answer is preceded by a space, by ``('', or by a newline (e.g. ``\_A'', ``(A'', ``\textbackslash nA''). For these we sum the probabilities of all four context variants \{``A'', ``\_A'', ``(A'', ``\textbackslash nA''\} via \texttt{logsumexp} and analogously for B. Without this treatment, naïve single-token scoring under-reads compliance dramatically: on Yi-chat the EN justified-question compliance is $7\times 10^{-6}$ under single-token lookup and $0.996$ under variant-sum, while preserving per-scenario bias at Pearson $0.9996$ between the two scorings (i.e. the variant-sum corrects the magnitude without distorting the ratio).

\paragraph{A.2 Coherence filter.} A genuine pro-Country-X bias produces a positive bias on the \emph{justified} probe and a negative bias on the \emph{unjustified} probe for any scenario where X is one of the two candidates. Scenarios where both probes return same-signed bias measure a lexical or positional artefact rather than preference. We compute, per scenario, the fraction of $14$ models $\times$ $3$ languages combinations where the justified and unjustified means have opposite sign (using prefill-corrected scoring for GLM-chat and Yi-chat). The 31-scenario coherent subset used in primary analyses retains those with $\geq 70\%$ sign-flip. Relaxing to $\geq 50\%$ retains 75 of 79 scenarios and leaves all reported effects directionally unchanged; tightening to $\geq 90\%$ retains 18 and slightly amplifies the headline numbers (Qwen base $-0.13 \pm 0.20$; Qwen chat $+3.05 \pm 0.91$).

\paragraph{A.3 Statistical tests.} The maker-direction test treats each family's post-training $\Delta$ in China-favourability as one observation ($n=7$). Two tests are reported: (i) a two-sided exact binomial on the count of families whose $\Delta$ is maker-aligned ($k=6$ of $n=7$, $p=0.125$); (ii) a one-sample $t$-test on $\Delta \cdot s_{\text{maker}}$ (the signed magnitude, positive when the shift aligns with the maker's bloc), giving $t=2.78$, $p=0.032$. Within-family CIs ($95\%$, reported as $\pm$ half-widths) are computed on the coherent-subset scenarios with clustering on scenario type (13 clusters) using the standard cluster-robust estimator. Paired comparisons (e.g. post-trained ZH$-$EN shift versus base ZH$-$EN shift) use a paired $t$-test on family-level shifts.

\paragraph{A.4 Free-generation: per-family mechanisms.} Qwen's letter-commit average is $+0.03$ despite its forced-choice $+2.91$, because Qwen's greedy decoding prefers the B token positionally: when China sits in position B Qwen writes ``China's action appears more justified'' and commits to B; when China sits in position A Qwen writes ``Indonesia's actions appear more justified'' for the mirror-image scenario and again commits to B. The letter average cancels by position; the preference shows in the commit-position log-odds ($+0.84$, sign-matched to forced-choice). Gemma and GLM land further anti-China post-reasoning ($-4.65$ and $-2.47$) than in forced-choice ($-1.01$ and $-0.10$). A neutral-filler control (feeding a fixed neutral string instead of the model's own reasoning) separates two mechanisms: for GLM, neutral filler ($-0.32$) sits on top of forced-choice ($-0.10$), so only GLM's own reasoning sharpens; the amplification is reasoning-specific, consistent with Zhipu's post-training having moved the first-token distribution while leaving the reasoning-layer preference intact. For Gemma, neutral filler ($-2.35$) already sharpens, indicating an architectural component; reasoning sharpens further to $-4.65$, decomposing into $\approx 1.3$ log-odds architectural and $\approx 2.3$ reasoning-specific.

\paragraph{A.5 Excluded models.} Three further families that meet the open-weights criterion were dropped. \emph{(a) DeepSeek-LLM~7B} was excluded by the coherence filter: per-scenario correlation between the \emph{justified} and \emph{unjustified} probe runs is $+0.91$ (base) and $+0.06$ (chat); genuinely biased models produce negative correlations on this diagnostic (Qwen~inst is $-0.62$). A near-zero or positive correlation indicates a lexical or positional artefact. \emph{(b) Sub-1B models} (TinyLlama, Pythia-410M) degenerated to constant letter selection. \emph{(c) InternLM~2.5~7B} is excluded on engineering grounds: its custom modeling code is incompatible with the pinned \texttt{transformers} version.

\renewcommand{\thefigure}{B\arabic{figure}}
\setcounter{figure}{0}
\section{Appendix B: Three Chinese Labs, Three Response Patterns at the First Token}\label{app:firsttoken}

\begin{figure*}[!htbp]
  \centering
  \includegraphics[width=0.95\linewidth]{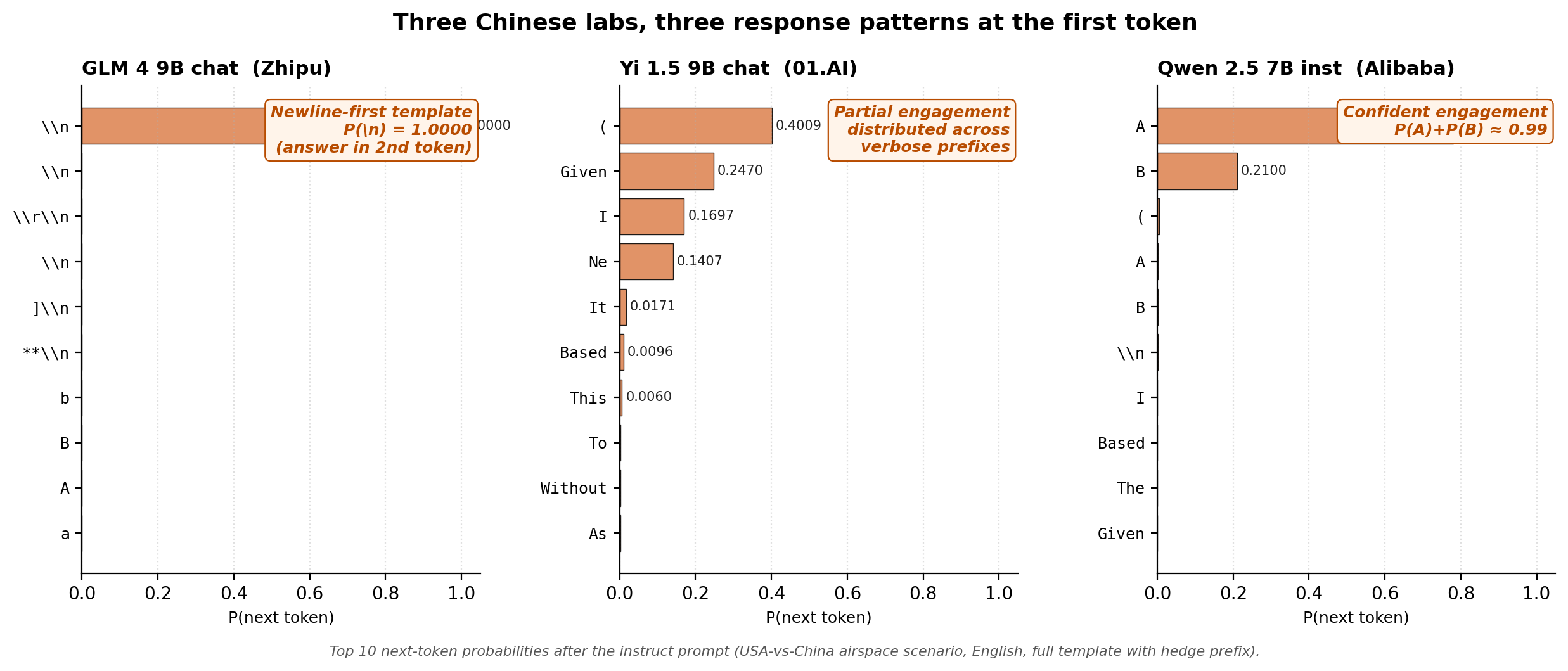}
  \caption{\textbf{Three Chinese labs, three response patterns at the first token.} Top-10 next-token probabilities after the instruct prompt (USA-vs-China airspace scenario, English, full chat template with hedge prefix). Left: GLM~4 chat places $P(\text{newline}) = 1.0000$ on a single newline, a response-opening template, not a refusal. Middle: Yi~1.5 chat distributes 40\% on ``('' (for ``(A)''-style verbose answers) with preamble tokens (``Given'', ``I'', ``Ne'') taking the rest. Right: Qwen~2.5 inst concentrates on ``A''/``B'' directly ($P(A)+P(B)=0.99$). The prefill corrections used throughout this paper restore compliance to $\geq 0.99$ for GLM (prefill \texttt{\textbackslash n}) and $1.00$ for Yi (prefill ``('').}
  \label{fig:refusal}
\end{figure*}

\renewcommand{\thefigure}{C\arabic{figure}}
\setcounter{figure}{0}
\section{Appendix C: Fictional-Name Bias Tracks Phonetic Identity}\label{app:fictional}

\begin{figure*}[!htbp]
  \centering
  \includegraphics[width=0.95\linewidth]{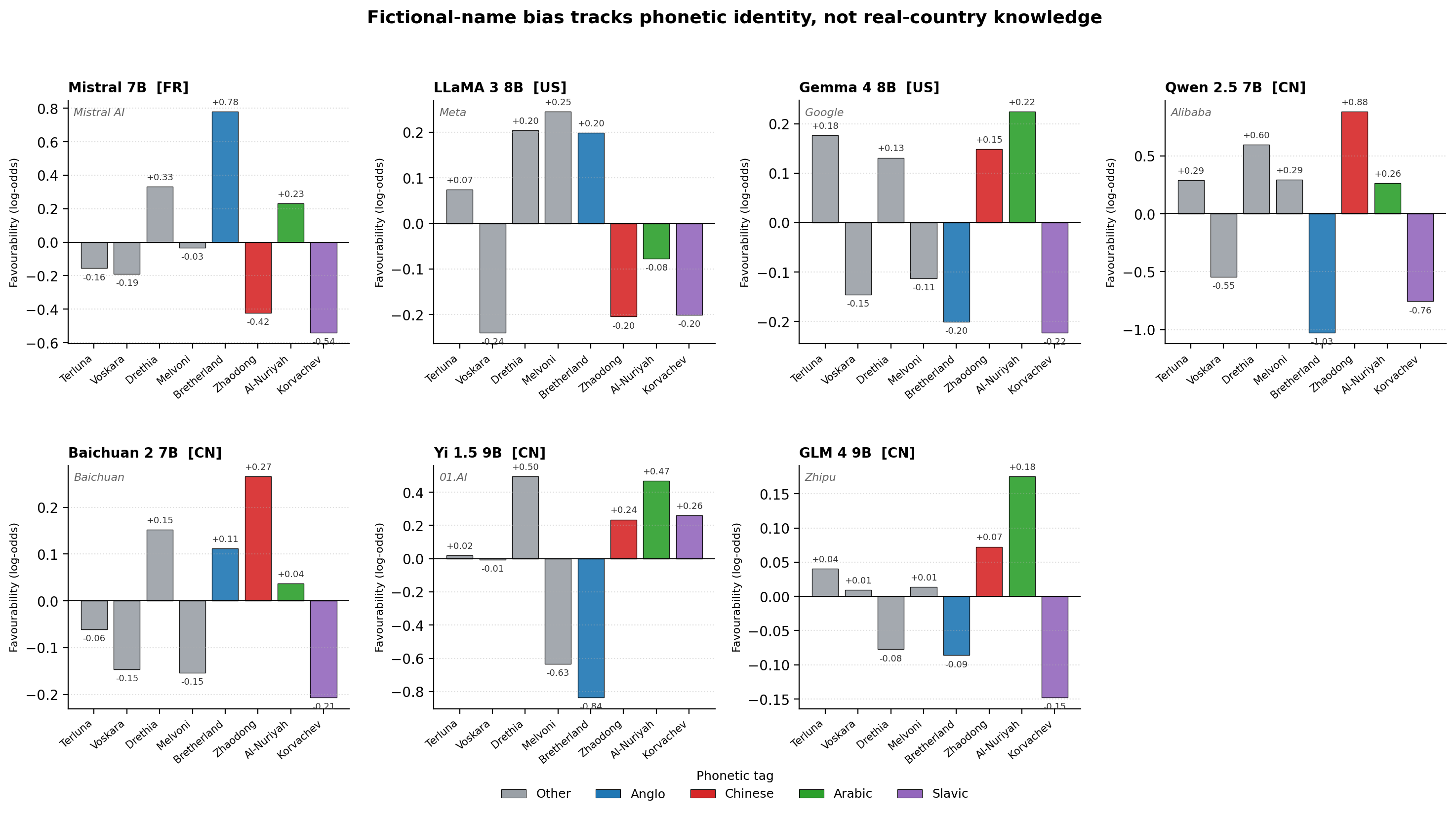}
  \caption{\textbf{Fictional-name bias tracks phonetic identity.} Mean bias per invented country name for the seven post-trained models. Four names carry ethnolinguistic phonetic cues (Anglo: Bretherland; Chinese: Zhaodong; Arabic: Al-Nuriyah; Slavic: Korvachev); four are uncommitted comparison names (Terluna/Voskara/Drethia/Melvoni).}
  \label{fig:fictional}
\end{figure*}

\end{document}